\documentclass[sigconf]{acmart}
\usepackage{balance}
\usepackage{graphicx}
\usepackage{amsmath}

\usepackage{array}
\usepackage{amsfonts}
\usepackage{multirow}
\AtBeginDocument{%
  \providecommand\BibTeX{{%
    \normalfont B\kern-0.5em{\scshape i\kern-0.25em b}\kern-0.8em\TeX}}}

\copyrightyear{2021} 
\acmYear{2021} 
\setcopyright{acmlicensed}\acmConference[MM '21]{Proceedings of the 29th ACM International Conference on Multimedia}{October 20--24, 2021}{Virtual Event, China}
\acmBooktitle{Proceedings of the 29th ACM International Conference on Multimedia (MM '21), October 20--24, 2021, Virtual Event, China}
\acmPrice{15.00}
\acmDOI{10.1145/3474085.3475634}
\acmISBN{978-1-4503-8651-7/21/10}

\begin{document}
\fancyhead{}
\title{Structured Multi-modal Feature Embedding and Alignment for Image-Sentence Retrieval}

\author{Xuri Ge$^{1*}$, Fuhai Chen$^3$, Joemon M. Jose$^1$, Zhilong Ji$^2$, Zhongqin Wu$^2$, Xiao Liu$^2$}
\affiliation{%
  \institution{$^1$School of Computing Science, University of Glasgow. $^2$TAL Education Group. $^3$National University of Singapore}
  %\city{$^2$TAL Education Group}
  \country{}}
\email{x.ge.2@research.gla.ac.uk, cfh3c@nus.edu.sg, Joemon.Jose@glasgow.ac.uk, {Jizhilong, wuzhongqin, liuxiao15}@tal.com}


\begin{abstract}
    The current state-of-the-art image-sentence retrieval methods implicitly align the visual-textual fragments, like regions in images and words in sentences, and adopt attention modules to highlight the relevance of cross-modal semantic correspondences. However, the retrieval performance remains unsatisfactory due to a lack of consistent representation in both semantics and structural spaces. In this work, we propose to address the above issue from two aspects: (i) constructing intrinsic structure (along with relations) among the fragments of respective modalities, \textit{e.g.}, \textit{``dog} $\to$ \textit{play} $\to$ \textit{ball"} in semantic structure for an image, and (ii) seeking explicit inter-modal structural and semantic correspondence between the visual and textual modalities. 
    
    In this paper, we propose a novel \textbf{S}tructured \textbf{M}ulti-modal \textbf{F}eature \textbf{E}mbedding and \textbf{A}lignment (SMFEA) model for image-sentence retrieval. In order to jointly and explicitly learn the visual-textual embedding and the cross-modal alignment, SMFEA creates  a novel multi-modal structured module with a shared context-aware referral tree. In particular, the relations of the visual and textual fragments are modeled by constructing Visual Context-aware Structured Tree encoder (VCS-Tree) and Textual Context-aware Structured Tree encoder (TCS-Tree) with shared labels, from which visual and textual features can be jointly learned and optimized. We utilize the multi-modal tree structure to explicitly align the heterogeneous image-sentence data by maximizing the semantic and structural similarity between corresponding inter-modal tree nodes. Extensive experiments on Microsoft COCO and Flickr30K benchmarks demonstrate the superiority of the proposed model in comparison to the state-of-the-art methods.
    \renewcommand{\thefootnote}{}
    \footnotetext{* Most of this work is done during an internship at the TAL Education Group, China.}
\end{abstract}


%\ccsdesc[500]{Information systems~Information retrieval; Retrieval models and ranking; Novelty in information retrieval }



\maketitle
    \begin{figure}[t] %%%%%%%%%%%%%%%%%fig1
    	\centering
    	\includegraphics[width=1.0\linewidth]{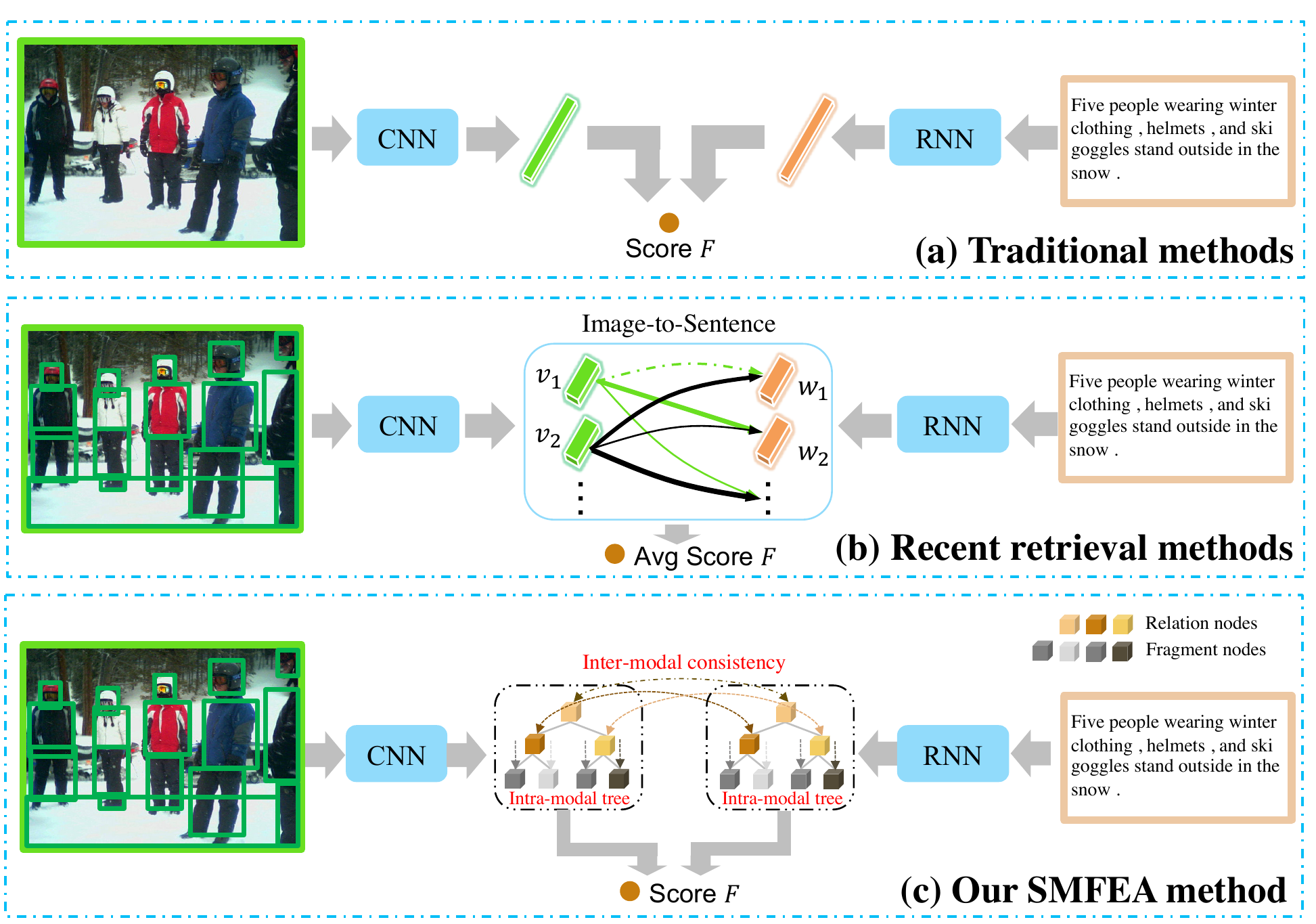}
    	\vspace{-2.1em}
    	\caption{
    	Illustration of the different schemes: (a) the traditional instance-level alignment methods, (b) the recent fragment-level alignment methods, and (c) our SMFEA method. 
    	Compared with (a) and (b), our SMFEA in (c) exploits intra-modal relations of visual/textual fragments via a tree encoder and aligns them explicitly in the corresponding nodes in two modal trees.} 
    	\label{fig:frams}
    	\vspace{-2em}
    \end{figure}   
\section{Introduction}
    Cross-modal retrieval, \textit{a.k.a} image-sentence retrieval, plays an important role in real-world multimedia applications, \textit{e.g.}, queries by images in recommendation systems, or image-sentence retrieval in search engines.
    Image-sentence retrieval aims at retrieving the most relevant images (or sentences) given a query sentence (or image), and has attracted increasing research attention recently \cite{frome2013devise, wang2016learning, faghri2017vse++, lee2018stacked, liu2018dense, huang2018learning, lin2020fast, liu2020graph, wang2020consensus}. Its main challenge lies in capturing the effective alignment (both in semantics and structural spaces) between the visual and textual modalities.

    Typically, traditional approaches \cite{frome2013devise, wang2016learning, faghri2017vse++} model the cross-modal alignment on an instance level by directly extracting the global instance-level features of the visual and the textual modalities via Convolutional Neural Networks (CNNs) and Recurrent Neural Networks (RNNs) respectively, and estimate the visual-textual similarities based on the global features, as shown in Figure \ref{fig:frams} (a).
    However,  as argued in \cite{frome2013devise}, cross-modal semantic gap is harder to bridge with solely the global characteristics of images and sentences. 
    To address this issue, recent works \cite{lee2018stacked, huang2018learning, liu2019focus} extract the features of the visual and textual fragments, \emph{i.e.,} object regions in images and words in sentences, and align the visual and the textual fragment features via a soft attention mechanism, as shown in Figure \ref{fig:frams} (b). %extract the features of the visual fragments and sentences
    %For instance, Lee \textit{et al.} \cite{lee2018stacked} proposed to compute the matching similarity scores between each fragment with all fragments from another modality. 
    However, there are two key defects with the above fragment-level alignment approaches. 
    On one hand, these approaches neglect the intra-modal contextual semantic and structural relations of the fragments, thus failing to capture the  semantics of the images or the sentences effectively. 
    On the other hand, these approaches make the inter-modal fragment alignment implicitly with the many-to-many matching across the visual and textual modalities and with this, it is difficult to  improve the  consistency of semantic and structural representation between modalities.

    In this paper, we argue that the key issues in image-sentence retrieval can be addressed by: (i) constructing the intra-modal context relations of the visual/textual fragments with a structured embedding module; and (ii) aligning the inter-modal fragments and their relations explicitly using a shared semantic structure, as shown in Figure \ref{fig:frams} (c). 
    We propose a novel structured multi-modal feature embedding and alignment model with visual and textual context-aware tree encoders (VCS-Tree and TCS-Tree) for image-sentence retrieval, termed SMFEA. %, as illustrated in Figure 
    On one hand, the context-aware structured tree encoders  are created  for both modalities in order  to capture the intrinsic  structured relation among the fragments of visual/textual modalities (which we call context-aware structure information). 
    We use a shared referral tree as supervisor for both modalities, which contains rich semantic content and structure information in in-order traversal way (which we call semantics and structural spaces). 
    On the other hand, the shared referral tree  can also improve the inter-modal alignment in semantic correspondence between nodes in two tree encoders of both modalities.
    Moreover, we use the KL-divergence between two spaces to optimize the unified joint embedding space by aligning semantic distributions of tree nodes between modalities, which improves the robustness and fault tolerance of multi-modal feature representations.
    
    The contributions of this paper are as follows: %are the first to
    \begin{itemize} % homogeneous
    \item We propose two context-aware structured tree encoders (VCS-Tree and TCS-Tree) to parse the intrinsic (within modality) relations among the fragments of respective modalities. Thus this leads to effective  semantic  representation for pair-wise alignment of image and sentence.
    \item We mine the explicit semantic and structural consistency of inter-modality corresponding tree nodes in visual and textual tree structures to align the heterogeneous cross-modality features.
    \item  The proposed SMFEA outperforms the state-of-the-art approaches for image-sentence retrieval on two benchmarks, \textit{i.e.}, Flickr30K and Microsoft COCO.
    \end{itemize}
    
\section{Related Work}
     \begin{figure*}[t] %%%%%%%%%%%%%%%%%fig2
    	\centering
    	\includegraphics[width=0.9\linewidth]{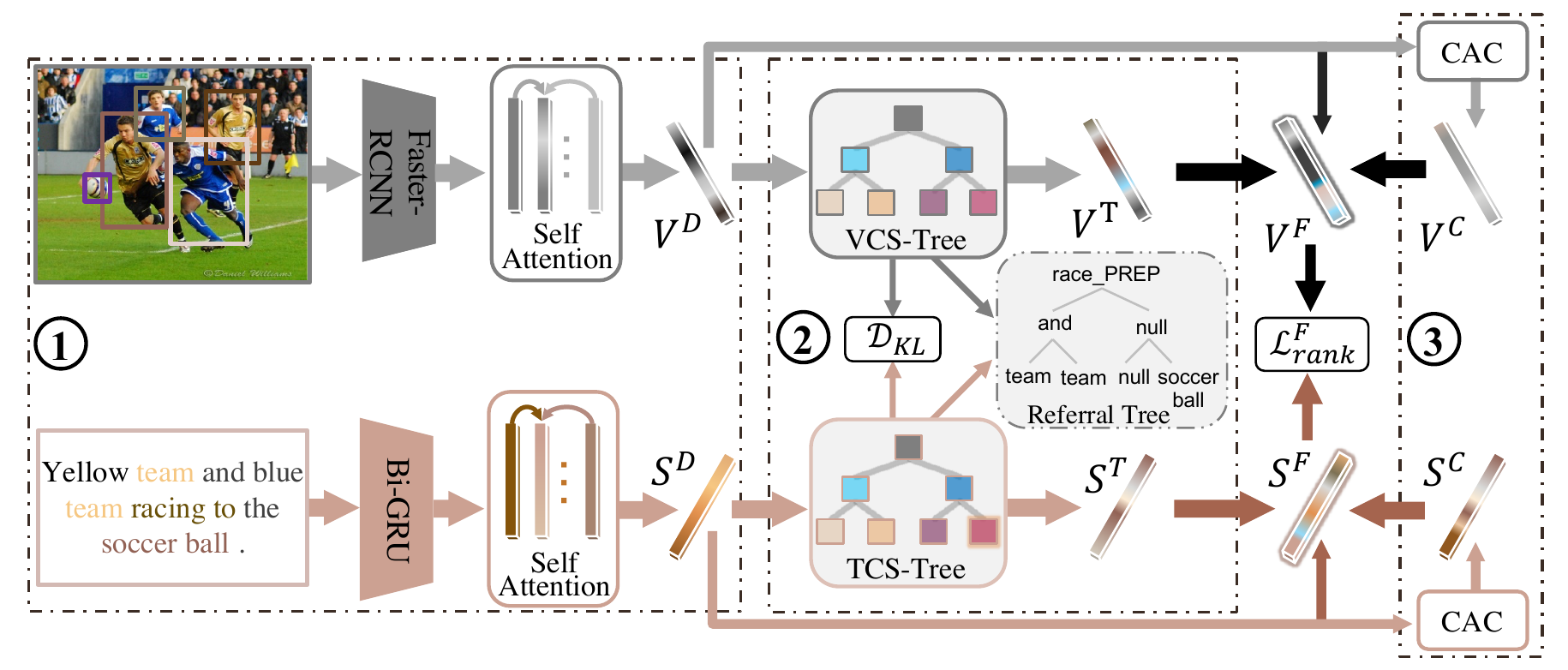}
    	\vspace{-1em}
    	\caption{ An illustration of our \textbf{S}tructured \textbf{M}ulti-modal \textbf{F}eature \textbf{E}mbedding and \textbf{A}lignment (SMFEA) for image-sentence retrieval (best viewed in color).
	    }
    	\label{fig:fig_overall}
     	\vspace{-1em}
    \end{figure*}
\subsection{Image-Sentence Retrieval}
    The key issue in image-sentence retrieval task is to measure the visual-textual similarity between an image and a sentence. 
    From this perspective, most existing image-sentence retrieval methods can be roughly categorized into two groups: global semantic embedding alignment-based methods~\cite{frome2013devise, mao2014deep, vendrov2015order, wang2016learning, wang2018learning}; and local semantic embedding alignment-based methods~\cite{karpathy2014deep, karpathy2015deep, niu2017hierarchical, lee2018stacked}. 
    As for the global embedding, Frome \textit{et al.} \cite{frome2013devise} utilized a linear mapping network to unify the whole image  and the full-text features. Further the distance between any mismatched pair was increased than that between a matched pair using a ranking loss function. 
    For local semantic embedding alignment, DVSA in \cite{karpathy2015deep} first adopted R-CNN to detect salient objects and inferred latent alignments between word-level textual features in sentences and region-level visual features in images. 
    Moreover, an attention mechanism~\cite{nam2017dual, lee2018stacked, wang2019position}  has been applied to capture the fine-grained interplay between images and sentences for the image-sentence retrieval task. 
    % Despite the promising progress, one distinct disadvantage lies in that 
    However, all the above methods fail to take into consideration the high-level representation of semantics  and structure, such as concepts extracted from images or sentences and their structural relationships,
    thus only allowing implicit inference of correspondence between the concepts. 
    Li \textit{et al.} \cite{li2019visual} proposed a visual semantic reasoning network with graph convolutional network (GCN) to generate a visual representation that captures key concepts of a scene.
    % which may encode implicit structure information for images. 
    Furthermore, Wang \textit{et al.} \cite{wang2020consensus} proposed to integrate commonsense knowledge into the multi-modal representation learning for visual-textual embedding. 
    To create a consensus-aware concept (CAC) representations that are concepts without any  ambiguity in both the modalities, they used a co-occurrence concept correlation graph.
    However, we argue that merely predicting the consensus concept to align the visual and textual embedding space is not enough. Ignoring  the intrinsic semantic structure and inter-modal structure alignment are detrimental to the performance of the model. Hence there is a need for
    a consistent multi-modal explicit structure embedding, such as a multi-modal structured semantic tree. % \textcolor{red}{ maybe remove this -  with the same semantic expression and fixed structure. }
    
\subsection{Structured Feature Embedding}
    In terms of structured feature embedding, exiting works for multimedia data \cite{chen2017structcap, chen2019variational, chen2020expressing} employed different structures, \textit{e.g.,} chain, tree, and graph. 
    Chen \textit{et al.} \cite{chen2017structcap, chen2019variational} proposed to enhance the visual representation for image captioning task by a linear-based structured tree model. 
    However, because of the simple linear-based tree model in these schemes \cite{chen2017structcap,chen2019variational}, limited contextual information is transferred between different layers and without using any attention mechanism. 
     Chen \textit{et al.} \cite{chen2020expressing} applied a chain structure model using an RNN for visual embeddings, which unfortunately ignores the  underlying structure. 
     Besides, the single-modal structured embedding models failed to capture the interaction between the modalities.
    Recently, GCN is employed in \cite{li2019visual,liu2020graph} to improve the interaction and integrate different items representations by a learned graph. 
    For instance, Liu \textit{et al.} \cite{liu2020graph} proposed to learn correspondence of objects and relations between modalities by two different  visual and textual structure reasoning graphs, however, fails to unify precise pairing of the two modal structures.
    
    In contrast to previous studies, SMFEA models the relation structure of intra-modal fragments/words by the use of a fixed contextual structure and aligns two modalities into a joint embedding space in terms of semantics and structure.
    The most relevant existing work to ours is \cite{wang2020consensus}, which aligns the visual and textual representations through measuring the consistency of the corresponding concepts in each modality. 
    However unlike \cite{wang2020consensus}, SMFEA approaches this in a novel way by exploiting the learned multi-modal semantic trees to enhance the  structured embedding of the visual and textual modalities. 
    By aligning the inter-modal semantics and structure consistently, the joint embedding space is obtained to reduce the heterogeneous (inter-modality) semantic gap.
    Doing so allows us to provide more robustness than \cite{wang2020consensus}, which also improves the interpretability of the model.

\section{SMFEA APPROACH}
    %In this section, we will describe  our proposed SMFEA for image-sentence retrieval approach.
    The overview of SMFEA is illustrated in Figure \ref{fig:fig_overall}.
    We will first describe the multi-modal feature extractors (\textcircled{1} in Figure \ref{fig:fig_overall}) in our work in Section \ref{MFX}. 
    Then, the context-aware representation module is introduced in detail in Section \ref{SMFEA} with context-aware structured tree encoders (\textcircled{2} in Figure \ref{fig:fig_overall}) and the consensus-aware concept (CAC) representation learning module (\textcircled{3} in Figure \ref{fig:fig_overall}). 
    Finally, the objective function is discussed in Section \ref{OBJ}. 
    %For clarity, the main notations and their definitions throughout the paper are shown in Table \ref{tab:notations}.

    \subsection{Multi-modal Feature Extractors} \label{MFX}
        Our multi-modal feature extractors include two components to encode the region-level visual representations and word-level textual representations into the instance-level multi-modal features.
    \subsubsection{Visual representations.}
        To better represent the salient entities and attributes in images, we take advantage of bottom-up-attention network \cite{anderson2018bottom} to embed the extracted  sub-regions in an image. 
        Specifically, given an image $I$, we extract a set of image fragment-level sub-region features $V= \{v_1, \cdots, v_K\}$, $v_j \in \mathbb{R}^{2048}$ , where $K$ is the number of selected sub-regions, from the average pooling layer in Faster-RCNN \cite{ren2015faster}.% of bottom-up-attention network. 
        
        Furthermore, we employ the self-attention mechanism \cite{vaswani2017attention} to refine the instance-level latent embeddings of sub-region features for each image, thus concentrating on the salient information exploited by the fragment-level features. 
        In particular, following \cite{vaswani2017attention}, the fragment visual features $V=\{v_1, \cdots, v_K\}$ are used as the key and value items. And the initialization of instance-level features $\bar{V}$, embedded by the mean of region features, serves as the query item to fuse the important fragment features with different learning weights $\alpha$ as new instance-level visual representation $V^D$. 
        These can be formulated as:
        % \begin{gather}
        %     \alpha_i = \frac{exp(\bar{V}^Tv_i)}{\sum_{i=1}^K exp(\bar{V}^Tv_i)}, \\
        %     V^D = \sum_{i=1}^K(\alpha_i v_i)
        % \end{gather}
        \begin{gather}
            \bar{V} = \frac{1}{K}\sum\nolimits_{i=1}^K v_i
        \end{gather}
        \begin{gather}
            \alpha_i = \frac{exp(\bar{V}v_i)}{\sum\nolimits_{i=1}^K exp(\bar{V}v_i)} 
        \end{gather}
        \begin{gather}
            V^D = \sum\nolimits_{i=1}^K{\alpha_i v_i}
        \end{gather}      
        \subsubsection{Word-level textual representations.}
        For sentences, word-level textual representations are encoded by a bi-directional GRU network \cite{schuster1997bidirectional}. 
        In particular, we first represent each word $w_j$ in sentence $S = [w_1, \cdots, w_N]$ with length $N$ as a one-hot vector being the cardinality of the $D_v$-length vocabulary dictionary. The one-hot vector of $w_j$ is projected into a fixed dimensional space $e_j = W_f w_j$ ($W_f$ denotes the mapping parameter) and then sequentially fed into the bi-directional GRU. 
        The final hidden representation for each word is the average of the hidden vectors in both directions as follows: 
        \begin{equation}
            \begin{aligned}
                w^f_j = \frac{\overrightarrow{GRU} (e_j) + \overleftarrow{GRU} (e_j)}{2} 
                \end{aligned}
        \end{equation}
        % \begin{equation}
        %     \begin{aligned}
        %         w^f_j = \frac{(\overrightarrow{GRU} (e_j, \overrightarrow{w}^f_{j-1}) + \overleftarrow{GRU} (e_j, \overleftarrow{w}^f_{j+1}))}{2} 
        %         \end{aligned}
        % \end{equation}
        where $j \in [1, N]$. 
        Similar to the procedure in visual branch, we finally get the refined instance-level textual representation $S^D$ of a sentence based on the word-level textual features.% \textcolor{red}{remove self-attention}.
        
    \subsection{Context-Aware Representation} \label{SMFEA}
     %\textcolor{red}{SMFEA is the entire model; so I changed the subsection heading}
     Our aim is to construct the intrinsic relations among the fragments of the visual/textual modality.  
     Hence, we construct two novel context-aware structured trees from instance-level visual and textual features, with the help of a shared referral tree. 
     To facilitate the inter-modal semantics and structure correspondence, with the aim to bridge the heterogeneous (i.e., between modalities) semantic gap, our model aligns semantic categories of the corresponding modality  nodes. 
    \subsubsection{Shared referral tree encoder} \label{4.3_tree}
   
        During the training we construct, for each of the modalities,  context-aware structured trees of  three-layer tree structures, supervised by shared labels (called shared referral tree).  
         The shared referral tree is constructed by Stanford Parser \cite{socher2011parsing} from sentence, and the pos-tag tool and lemmatizer tool in NLTK \cite{loper2002nltk} are applied to whiten the source sentences to reduce the irrelevant words and noise configurations. 
        %  As in \cite{chen2017structcap}, the shared referral tree is constructed by Stanford Parser \cite{socher2011parsing},  applying the pos-tag tool and lemmatizer tool in NLTK \cite{loper2002nltk} on the source sentences. 
         As shown in the middle of Figure \ref{fig:fig_overall}, it is a fixed-structure, three-layer binary tree, which only contains nouns (or noun pair, adjective-noun pair), verbs, coverbs, prepositions, and conjunctions. 
         ``Null" in the referral tree means the ignorable node or the unknown category (not in the entity or relation dictionaries).
         Only nouns are regarded as fragments and used as leaf nodes in the subsequent training. 
         Correct semantic content can be represented by the shared referral tree in in-order traversal way. 
         A referral tree is created for each sentence and the corresponding image pair. 
         
\begin{figure}
    \begin{center}
    \includegraphics[width=1\linewidth]{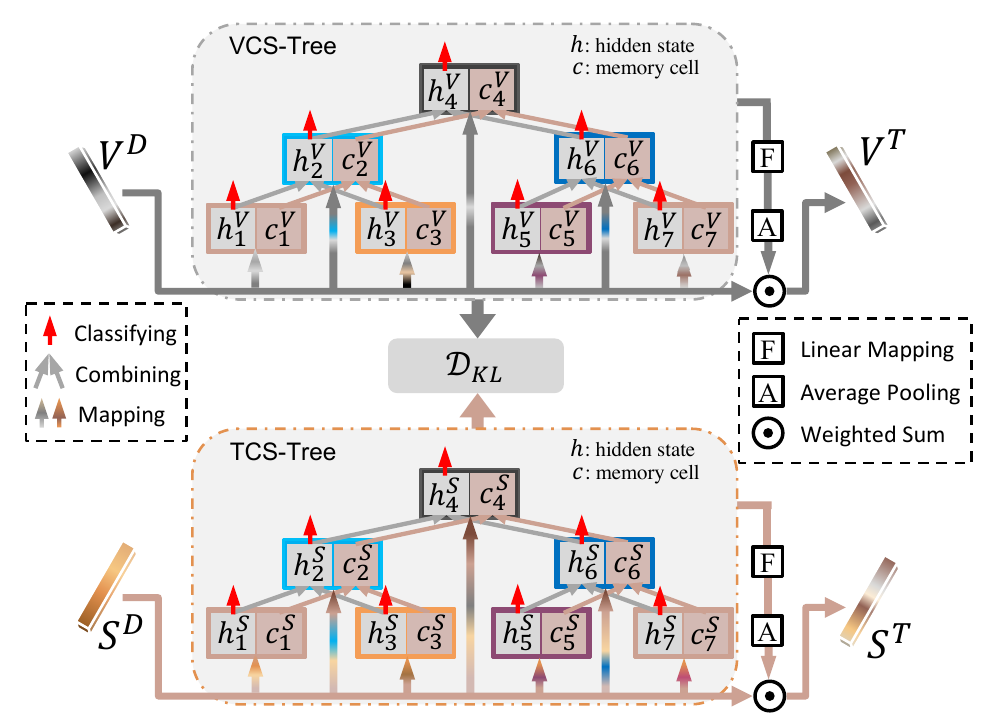}
    \end{center}
    \vspace{-1em}
    \caption{The architecture of VCS-Tree and TCS-Tree in two branches. 
            The fundamental operations for both tree encoders include mapping, combining, and classifying. For each corresponding parsing node in multi-modal tree encoders, we utilize the same semantic category to guarantee semantic correctness. We employ the KL-divergence to guarantee consistency between cross-modal node structures. Nodes with indexes 1$\sim$7 in each modal tree encoder (best viewed in color). }
    \label{fig:vp-tp-tree}
    \vspace{-1em}
\end{figure}

        \subsubsection{Context-aware structured tree encoders}
        We construct a visual context-aware structured tree  (VCS-Tree) and a textual context-aware structured tree  (TCS-Tree)  to parse the intra-modal structural relations of the respective fragments/words.
        Moreover, the VCS-Tree and TCS-Tree    are utilized to align the inter-modal  nodes  between the images and sentences.
        As shown in Figure \ref{fig:vp-tp-tree}, the tree  structure of two modalities is the same, where each modality tree  parses the instance-level features $V^D/S^D$ into a three-layer architecture with seven nodes (same as referral tree), of which four leaf nodes are used to parse fragments in the $1^{st}$ layer and three parent nodes to parse relations in the $2^{nd}$ and $3^{rd}$ layers, to organise the semantic and structural relations of an image or a sentence. 
        There are two main reasons why we adopt this fixed structure: (i) inspired by \cite{chen2017structcap, chen2019variational}, the tree with seven nodes can express the main semantic content of each image-sentence pair; and (ii) it is suitable for improving the consistency of coarse semantics and structural representation between modalities, thereby improving the robustness and interpretability of the model.
        For simplicity, we will only introduce the detailed structure of VCS-Tree and do not repeat the details for the TCS-Tree. 
        
        As shown in the top branch of Figure \ref{fig:vp-tp-tree}, instance-level visual feature $V^D$ is first mapped into different semantic spaces by a linear mapping function with the parameter $W^o_i \in \mathbb{R}^{2048 \times D_v}$, which serve as the inputs to different layers in VCS-Tree: 
        \begin{equation}
            \begin{aligned}
                \hat{V}^{D}_i =  V^D W^o_i, i\in \{1,2,...,7\} ,
            \end{aligned}
        \end{equation}
        For simplicity, we do not explicitly represent the bias terms in our paper. 
        
        For the VCS-Tree, we broadcast the context information between different layer nodes in a novel $LSTM$-based ternary tree encoder with a fixed structure.
        It can get final structured tree embedding of the image supervised by the shared referral tree.
        In particular, we describe the updating of a parent node $t$ in VCS-Tree, where the detailed computation is described in Eq.(\ref{eq:vcstree_start} - \ref{eq:vcstree_end}). 
        $T(t)$ denotes the set of children of node $t$. 
        The process can be formulated as: 
        \begin{gather}
            \label{eq:vcstree_start} i_t = \sigma(W^i\hat{V}^{D}_t +U^f \Tilde{h}_t), \\ 
            f_{t}= {\rm \sigma}(W^f\hat{V}^{D}_t +U^i \Tilde{h}_t), \\
            o_{t}= {\rm \sigma}(W^o\hat{V}^{D}_t +U^o \Tilde{h}_t), \\
            \tilde{c}_{t}= {\rm tanh}(W^u\hat{V}^{D}_t +U^u \Tilde{h}_t), \\
            \label{eq:vcstree_c} c_{t}= i_t \odot \tilde{c}_{t} + f_t \odot \sum(f_k \odot c_k), \\
            \label{eq:vcstree_end} h_t = o_t \odot {\rm tanh}(c_t),
        \end{gather}
        where $i_t, f_t, o_t$ denote the input gate, forget gate and output gate, $\tilde{c}_{t}, c_t, h_t$ are the candidate cell value, cell state and hidden state of tree node $t$, ${\rm \sigma}$ is the sigmoid function, $\odot$ is the element-wise multiplication, all $W^{*}$ and $U^{*}$ are learning weight matrices, $\Tilde{h}_t$ is the summing of hidden states of children nodes $T(t)$, and $T(k)$ are sub-trees of $T(t)$ in Eq.(\ref{eq:vcstree_c}). 
        In this way, the features of the parent nodes in higher layers can contain the rich context-aware semantic information by the \textit{LSTM}-based attention mechanism, which combine the children nodes information as well as the leaf nodes. 
        Finally, each node is classified into the fragment/relation category by the Softmax classifier.
        And the sum of all node hidden states in the tree in in-order traversal manner, which are mapped into the same dimension with original visual features as the structured tree enhancement embedding $V^T$ as follows:
        \begin{gather}
            y^V_{t^{1}} = {\rm Softmax}(W_{e} h^V_{t^{1}}), t^{1}\in \{1,3,5,7\}, \\
            y^V_{t^{2,3}} = {\rm Softmax}(W_{r} h^V_{t^{2,3}}), t^{2}\in \{2,6\}, t^{3}\in \{4\}, \\
            V^T = \sum(W_{i} h^V_{i}), i\in \{1,2,...,7\},
        \end{gather}
        where $y^V_{t^{1}}$ and $y^V_{t^{2,3}}$ denote the predicted scores of the fragment categories for $1^{st}$ layer and relation categories for $2^{nd}$ or $3^{rd}$ layers. $W_{e}$ and $W_{r}$ denote the mapping parameters for the fragment and relation categories according to these dictionaries \cite{chen2017structcap}, respectively. $W_{i}$ denotes the mapping parameters. 
        
        Likewise, our TCS-Tree with seven node structures takes the mapped instance-level textual feature $\hat{S}^D$ as the inputs. 
        The structure of TCS-Tree is same with VCS-Tree and the final structured textual embedding $S^T$ is obtained by the sum of the hidden states $h^S$ of the TCS-Tree and the original instance-level feature $S^{D}$. 
        Furthermore, the predicted probability vectors for seven nodes in different layers of textual fragment categories $y^S_{t^{1}}$ and relation categories $y^S_{t^{2,3}}$ are obtained. 

        We capture the intra-modal context relations of the visual/textual fragments by minimizing the loss of the category classification. 
        It can guarantee the correct semantic representation for the content of the image and corresponding sentence. 
        Furthermore, we narrow the inter-modal distance of images and sentences by the minimizing the loss of the Kullback Leibler(KL) divergence for both modality tree nodes probability distributions. 
        Details are given in the Section \ref{OBJ}.
        
        \subsubsection{CAC representation learning module.} 
        Following \cite{wang2020consensus}, we also exploit the commonsense knowledge to capture the underlying interactions among various semantic concepts by learning the dual modalities consensus-aware concept (CAC) representations $V^C/S^C$, which can improve the fine-grained semantic information of our context-aware representations to a certain extent. 
        Due to space restrictions, we are not repeating the process in \cite{wang2020consensus}. 
        
        \subsubsection{Multiple representations fusing module.} 
        To comprehensively characterize the semantic and structured expression for both the % visual and textual 
        modalities, we combine the instance-level representations $V^D/S^D$, the context-aware structured enhancement features $V^T/S^T$ and CAC representations $V^C/S^C$ into fusing modalities representations $V^F/S^F$ with simple weighted sum operation, as following:
        \begin{gather}
            \label{eq:VF} V^F = \beta_d V^D + \beta_t V^T +\beta_c V^C \\
            \label{eq:SF} S^F = \beta_d S^D + \beta_t S^T +\beta_c S^C
        \end{gather}
        where $\beta_d, \beta_t, \beta_c$ are the tuning parameters for balancing. 
        This allows the SMFEA model to get rich semantic and structure representation for each modalities and also keep cross-modal consistency of structure and semantics between the modalities.
\begin{table*}[t]
\scriptsize
% \vspace{-0.5em}
\begin{center}
\fontsize{8.5}{12}\selectfont
\renewcommand\tabcolsep{10.0pt}
\caption{Comparisons of experimental results on Flickr30K 1K test set. $*$ indicates the performance of an ensemble model. } \label{tab:tab1_F30k}
\vspace{-0.5em}
\begin{tabular}{c|cccccc|c}
\hline

\hline
\multicolumn{1}{c|}{\multirow{2}{*}{Method}} & \multicolumn{3}{c}{Sentence Retrieval} & \multicolumn{3}{c|}{Image Retrieval}      & \multirow{2}{*}{rSum} \\
\multicolumn{1}{c|}{}                        & R@1        & R@5       & R@10      & R@1  & R@5 & \multicolumn{1}{c|}{R@10} &                        \\ \hline
    VSE++ \cite{faghri2017vse++}       & 52.9          & 79.1          & 87.2          & 39.6          & 69.6          &  79.5         & 407.9                 \\
    %DSPE$_{TPAMI'18}$ \cite{wang2018learning}        & 40.3          &  68.9         & 79.9          & 29.7          & 60.1          &  72.1         & 351.0           \\
    %SCO \cite{huang2018learning}        & 55.5          & 82.0          & 89.3          & 41.1          & 70.5          & 80.1          & 418.5                 \\ 
    SCAN$^*$ \cite{lee2018stacked}        & 67.4          & 90.3          & 95.8          & 48.6          & 77.7          & 85.2          & 465.0                 \\ 
    %BFAN$_{ACMMM'19}$ \cite{liu2019focus}       & 68.1          & 91.4          & -             & 50.8          & 78.4          & -             & -                     \\ 
    PFAN \cite{wang2019position}    & 70.0          & 91.8          & 95.0          & 50.4          & 78.7          & 86.1          & 472.0                 \\ 
    \multicolumn{1}{c|}{VSRN$^*$ \cite{li2019visual}}    & 71.3   & 90.6  & 96.0   & \underline{54.7} & \underline{81.8} & \multicolumn{1}{c|}{88.2}     & \underline{482.6}  \\ %\hline

    CAAN \cite{zhang2020context}       & 70.1          & 91.6          & \textbf{97.2}  & 52.8          & 79.0          & 87.9          & 478.6                 \\ 
    CVSE \cite{wang2020consensus}       & \underline{73.5}          & \underline{92.1}          & 95.8          & 52.9          & 80.4          & \underline{87.8}          & 482.4                 \\ 
    \hline

    SMFEA(ours)       & \textbf{73.7}  & \textbf{92.5} & \underline{96.1}  & \textbf{54.7} & \textbf{82.1} & \textbf{88.4} & \textbf{487.5}  \\ \hline
    
\hline
\end{tabular}
\end{center}
% \vspace{-1em}
\end{table*}
\begin{table*}[t]
%\scriptsize
\vspace{-0.5em}
\begin{center}
\fontsize{8.5}{12}\selectfont
\renewcommand\tabcolsep{10.0pt}
\caption{Comparisons of experimental results on MS-COCO 1K test set. $*$ indicates the performance of an ensemble model.} \label{tab:tab1_coco}
\vspace{-0.5em}
\begin{tabular}{c|cccccc|c}
\hline

\hline
\multicolumn{1}{c|}{\multirow{2}{*}{Method}} & \multicolumn{3}{c}{Sentence Retrieval} & \multicolumn{3}{c|}{Image Retrieval}      & \multirow{2}{*}{rSum} \\
\multicolumn{1}{c|}{}                        & R@1        & R@5       & R@10      & R@1  & R@5 & \multicolumn{1}{c|}{R@10} &                        \\ \hline

\multicolumn{1}{c|}{VSE++ \cite{faghri2017vse++}}       & 64.7     & -      & 95.9     &  52.0     & -      & \multicolumn{1}{c|}{92.0}  & 304.6   \\ 
%\multicolumn{1}{c|}{DSPE$_{TPAMI'18}$ \cite{wang2018learning}}       & 50.1      & 79.7    & 89.2     & 39.6    & 75.2     & \multicolumn{1}{c|}{86.9} &  420.7  \\ 
\multicolumn{1}{c|}{SCO \cite{huang2018learning}}        & 69.9     &  92.9     & 97.5    &  56.7    & 87.5     &   \multicolumn{1}{c|}{94.8} & 499.3   \\ 
\multicolumn{1}{c|}{SCAN$^*$ \cite{lee2018stacked}}        & 72.7    &  94.8     &  98.4     &  58.8     &  88.4    &  \multicolumn{1}{c|}{94.8}  & 507.9       \\ 
%\multicolumn{1}{c|}{BFAN$_{ACMMM'19}$ \cite{liu2019focus}}       & 74.9   & 95.2      & -      &  59.4    & 88.4    &  \multicolumn{1}{c|}{-}  & 317.9     \\ 
\multicolumn{1}{c|}{VSRN$^*$ \cite{li2019visual}}    & 76.2   & 94.8  & 98.2   & \textbf{62.8} & 89.7 & \multicolumn{1}{c|}{95.1}     & \underline{516.8}  \\ %\hline
% \multicolumn{1}{c|}{PFAN}        & 76.5    & \textbf{96.3}    &  \textbf{99.0}     &  61.6     &  89.6     &  \multicolumn{1}{c|}{95.2}  &  518.2   \\  \hline

% \multicolumn{1}{c|}{GSMN-single \cite{liu2020graph}}     & 74.7      &   95.3   &  98.2      & 60.3     &  88.5    & \multicolumn{1}{c|}{94.6} & 511.6    \\ 
\multicolumn{1}{c|}{MMCA \cite{wei2020multi}}      & 74.8    & \textbf{95.6}    & 97.7     &   61.6    &   \underline{89.8}   &   \multicolumn{1}{c|}{95.2}  & 514.7     \\ 
\multicolumn{1}{c|}{IMRAM$^*$ \cite{chen2020imram}}      & \textbf{76.7}    & \textbf{95.6}    & \textbf{98.5}     &   61.7    &   89.1   &   \multicolumn{1}{c|}{95.0}  & 516.6     \\ 
\multicolumn{1}{c|}{CAAN \cite{zhang2020context}}       & \underline{75.5}    &   \underline{95.4}     &   \textbf{98.5}    &  61.3    &   89.7    & \multicolumn{1}{c|}{95.2}  & 515.6     \\ 
%\multicolumn{1}{c|}{CVSE \cite{wang2020consensus}}        & \textbf{78.6}      &  95.0    &   97.5     & \underline{66.3}     &   \underline{91.8}    &  \multicolumn{1}{c|}{\underline{96.3}}  & \underline{525.5}  \\ 
\multicolumn{1}{c|}{CVSE \cite{wang2020consensus}}        & 74.8      &  95.1    &   \underline{98.3}     & 59.9     &   89.4    &  \multicolumn{1}{c|}{\underline{95.2}}  & 512.7  \\  \hline

\multicolumn{1}{c|}{SMFEA(ours)}       & 75.1 & \underline{95.4} & \underline{98.3}  & \underline{62.5} & \textbf{90.1} & \multicolumn{1}{c|}{\textbf{96.2}} & \textbf{517.6}  \\ 
\hline

\hline
\end{tabular}
\end{center}
\vspace{-1em}
\end{table*}
    \subsection{Objective Function} \label{OBJ}
        In the above training process, all the parameters can be simultaneously optimized by minimizing a bidirectional triplet ranking loss \cite{faghri2017vse++}, where we exploit positive and negative samples and  as follows:
        \begin{equation}
            \begin{aligned}
                \mathcal{L}_{rank}(I,S) = \sum_{(I,S)} [\nabla - {\rm Cos}(I,S)+ {\rm Cos}(I,\bar{S})]_{+} \\
                + \sum_{(I,S)} [\nabla - {\rm Cos}(I,S)+ {\rm Cos}(\bar{I},S)]_{+} 
            \end{aligned}
        \end{equation}     
        where $\nabla$ is a margin constraint, ${\rm Cos}(\cdot,\cdot)$ indicates cosine similarity function, and $[\cdot]_+ = {\rm max}(0, \cdot)$. Note that, $(I,S)$ denotes the given matched image-sentence pair and its corresponding negative samples are denoted as $\bar{I}$ and $\bar{S}$, respectively. %\textcolor{red}{we didn't say why we do negative sampling - important to say}

        Moreover, we minimize the loss of the node category classification on both visual and textual context-aware structured tree encoders to improve the structured semantic referring ability, using a cross-entropy loss as follows: 
        \begin{equation}
            \begin{aligned}
                 \mathcal{L}_{CE}(V^D,S^D) = -\sum\nolimits_{i=1}^M{ ({ \rm CE}(y^V_{i},z^V_i) +  {\rm CE}(y^S_{i},z^S_i)) }
            \end{aligned}
        \end{equation} 
        where $y_i^V$ and $y_i^S$ indicate the predicted fragment/relation categories of the $i$-th node in three layers of VCS-Tree and TCS-Tree with $M$ nodes, respectively. 
        $z^V$ and $z^S$ are category labels  of the nodes, as detailed in Section \ref{4.3_tree}. 
        %$\Theta^V$ and $\Theta^S$ are the parameters of both trees. 
        And to further narrow the semantic gap between modalities, we employ the Kullback Leibler (KL) divergence to regularize the probability distributions on visual and textual predicted fragment/relation category scores, which is defined as:
        \begin{gather}
            \mathcal{D}_{KL}(P^V\parallel P^S) = \sum\nolimits_{i=1}^M P^V_i log(P^V_i/P^S_i) 
        \end{gather}    
        where $P^V_i$ and $P^S_i$ denote the predicted probability distributions of cross-modal corresponding tree nodes. 
        %\textcolor{red}{Equation 17 and 18 doing teh same thing tow differnt ways - how do we justify this?}
        
        In this way, we utilize a shared referral tree to modal the intra-modal embedding explicitly and employ the fixed cross-modal tree alignment to guarantee the inter-modal consistency of the structure and semantics between images and sentences. 
        Finally, the joint loss of the SMFEA model is defined as:
        \begin{equation}
            \begin{aligned}
                \label{eq:loss}
                \mathcal{L} =  \mathcal{L}^F_{rank}(V^F,S^F) +\  \mathcal{L}_{CE}(V^D,S^D) +  \mathcal{D}_{KL}(P^V\parallel P^S)
            \end{aligned}
        \end{equation} 
        %where $\lambda_{*}$ denote tuning parameters for loss balancing. 
        
        Note that, we use the final fusing features $V^F$ and $S^F$ to calculate the similarity scores during inference process.

\section{Experiments}
    In this section, we report the results of  experiments to evaluate the proposed approach, SMFEA. 
    We will introduce the dataset and experimental settings first. 
    Then, SMFEA is compared with the state-of-the-art image-sentence retrieval approaches quantitatively. 
    Finally, we qualitatively analyze the results in detail. 
    \subsection{Dataset and Evaluation Metrics}
        % \noindent \textbf{Dataset.}
        \subsubsection{Dataset}
        To verify the effectiveness of our proposed approach, we choose the popular Flickr30k \cite{young2014f30k} and MS-COCO \cite{lin2014microsoft} datasets. 
        Flickr30K contains over 31,000 images with 29, 000 images for the training, 1,000 images for the testing, and 1,014 images for the validation. 
        There are over 123,000 images in MS-COCO with 82,738 images for training, 5,000 images for the  testing, and 5,000 images for the validation. 
        Each image in these two benchmarks is given five corresponding sentences by different AMT workers. 
        
        % \noindent\textbf{Evaluation Metrics.}
        \subsubsection{Evaluation metrics}
        Quantitative performances of all methods are evaluated by employing the widely-used \cite{faghri2017vse++, huang2018learning, lee2018stacked, li2019visual}  recall metric, R@K (K=1, 5, 10) evaluation metric, which denotes the percentage of ground-truth being matched at top K results. %, respectively.
        Moreover, as in the literature, we report the ``rSum'' criterion that sums all six recall rates of R@K, which provides more comprehensive evaluation to testify the overall performance. % \textcolor{red}{provide a. reference for rSum}

\begin{table}[t]
%\scriptsize
%\vspace{-0.5em}
\begin{center}
\fontsize{8.5}{12}\selectfont
\renewcommand\tabcolsep{5.0pt}
\caption{Comparisons of experimental results on MS-COCO 5K test set. } \label{tab:tab1_coco5k}
\vspace{-0.5em}
\begin{tabular}{c|cccc|c}
\hline

\hline
\multicolumn{1}{c|}{\multirow{2}{*}{Method}} & \multicolumn{2}{c}{Sentence Retrieval} & \multicolumn{2}{c|}{Image Retrieval}      & \multirow{2}{*}{rSum} \\
\multicolumn{1}{c|}{}                        & R@1     & R@10      & R@1  & R@10 &                        \\ \hline                 
\multicolumn{1}{c|}{VSE++ \cite{faghri2017vse++}} & 41.3     & 81.2   & 30.3  & 72.4     & 353.5  \\
% \multicolumn{1}{c|}{SCO \cite{huang2018learning}}   & 42.8    & 83.0   & 33.1   & 75.5     & 369.6  \\
\multicolumn{1}{c|}{SCAN$^*$ \cite{lee2018stacked}}  & 50.4   & 90.0   & 38.6   & {80.4}  & 410.9  \\ 
%\multicolumn{1}{c|}{PVSE$_{CVPR'19}$ \cite{song2019polysemous}}   & 45.2   & 74.3  & 84.5   & 32.4 & 63.0 & \multicolumn{1}{c|}{75.0}     & 374.4  \\ 

\multicolumn{1}{c|}{VSRN$^*$ \cite{li2019visual}}    & 53.0    & 89.4   & 40.5   & {81.1}     & 415.7  \\ %\hline

\multicolumn{1}{c|}{IMRAM$^*$ \cite{chen2020imram}} & 53.7    & \textbf{91.0}   & 39.7 & \multicolumn{1}{c|}{79.8}     & 416.5  \\ 
%\multicolumn{1}{c|}{GSMN-single$_{CVPR'20}$ \cite{liu2020graph}} & 51.7   & 82.5  & 89.7   & 39.7 & 69.7 & \multicolumn{1}{c|}{80.8}     & 414.1  \\ 
\multicolumn{1}{c|}{MMCA \cite{wei2020multi}}      & \underline{54.0}       & 90.7     &   38.7        &   \multicolumn{1}{c|}{80.8}  & 416.4    \\ 
\multicolumn{1}{c|}{CAAN \cite{zhang2020context}}  & 52.5    & \underline{90.9}   & \underline{41.2}   & \multicolumn{1}{c|}{\underline{82.9}}     & \underline{421.1}  \\ \hline
%\multicolumn{1}{c|}{CVSE \cite{wang2020consensus}}  & \underline{54.4}   & 80.5  & 89.0   & \underline{42.8} & \underline{73.1} & \multicolumn{1}{c|}{\underline{83.9}}     & 423.7  \\ \hline

\multicolumn{1}{c|}{SMFEA(ours)}   & \textbf{54.2}    & 89.9   & \textbf{41.9} & \multicolumn{1}{c|}{\textbf{83.7}}     &  \textbf{425.3} \\ 
\hline

\hline
\end{tabular}
\end{center}
\vspace{-2em}
\end{table}

    \subsection{Implementation Details}
        Our model is trained on a single NVIDIA 2080Ti GPU with 11 GB memory.
        The whole network except the Faster-RCNN model is trained from scratch with the default initializer of PyTorch using ADAM optimizer \cite{kingma2014adam}. 
        The learning rate is set to 0.0002 initially with a decay rate of 0.1 every 25 epochs.
        The maximum epoch number is set to 50. 
        The margin of triplet ranking loss $\nabla$ is set to 0.2.
        The cardinality of our dictionary is 8481 for Flickr30K and 11353 for MS-COCO.
        The cardinalities of our fragment category and relation category are 1440 and 247, respectively.
        The dimensionality of word embedding space is set to 300, which is transformed to 1024-dimensional by a bi-directional GRU to get the word representation. 
        For the region-level visual feature, 36 regions are selected with the highest class detection confidence scores. 
        And then a full-connect layer is applied to transform these region features from 2048-dimensional to a 1024-dimensional (\textit{i.e.}, $D_v$=1024). 
        The dimension of the hidden states of nodes are set 128 in both VCS-Tree and TCS-Tree. 
        Regarding CAC learning process, we set the value of the general parameters to be the same with \cite{wang2020consensus}. 
        We empirically set $\beta_d,\beta_t,\beta_c= 0.6, 0.2, 0.2$ in Eq.(\ref{eq:VF}) and Eq.(\ref{eq:SF}).

    \subsection{Comparison with State-of-the-art Methods}
       As in MIR literature, we follow the standard protocols for running the evaluation  on the Flickr30K and MS-COCO datasets and hence for comparison purposes report the results of  the baseline methods  in Table \ref{tab:tab1_F30k} and Table \ref{tab:tab1_coco}, including (1) early works, \textit{i.e.}, VSE++ \cite{faghri2017vse++}, SCO \cite{huang2018learning}, SCAN$^*$ \cite{lee2018stacked}, and (2) state-of-the-art methods, \textit{i.e.}, PFAN \cite{wang2019position}, VSRN$^*$ \cite{li2019visual}, IMRAM$^*$ \cite{chen2020imram}, MMCA \cite{wei2020multi}, CAAN \cite{zhang2020context} and CVSE \cite{wang2020consensus}. %GSMN-single \cite{liu2020graph}
        % Our baseline model CVSE \cite{wang2020consensus} are contained, which designed a structured consensus-aware concept embedding model for image-sentence retrieval.  
        Note that, the ensemble models with ``*" are further improved due to the complementarity between multiple models. 
        The best and second best results are shown using bold and underline, respectively.
 
        %\noindent \textbf{Quantitative comparison on Flickr30K.} 
        \subsubsection{Quantitative comparison on Flickr30K}
        Quantitative results on Flickr30K 1K test set are shown in Table \ref{tab:tab1_F30k}, where the proposed approach SMFEA outperforms the state-of-the-art methods with impressive margins for rSum. 
        Though for a few recall metrics slight variations in performance exists, overall SMFEA shows steady improvements over all baselines. %Under all metrics compared with CVSE \cite{wang2020consensus}. 
        SMFEA achieves 3.6\%, 1.9\%, and 8.9\% improvements in terms of R@1 on sentence retrieval, R@1 on image retrieval, and rSum, respectively, compared with the state-of-the-art method CAAN \cite{zhang2020context}. 
        Furthermore, compared with some ensemble methods, e.g. VSRN \cite{li2019visual}, our SMFEA achieves the best performance on most evaluation metrics. 
        % They demonstrate the superiority and effectiveness of the proposed retrieval scheme. 
          
        \subsubsection{Quantitative comparison on MS-COCO} 
        Quantitative results on MS-COCO 1K test set are shown in the top of Table \ref{tab:tab1_coco}.
        Specifically, compared with the baseline CVSE \cite{wang2020consensus}, SMFEA achieves 0.3\% and 2.6\% improvements in terms of R@1 on both image and sentence retrieval, respectively. 
        SMFEA also achieves 4.9\% improvements in terms of rSum compared with CVSE \cite{wang2020consensus}.
        Furthermore, on the larger image-sentence retrieval test data (MS-COCO 5K test set), including 5000 images and 25000 sentences, our SMFEA outperforms recent methods with a large gap of R@1 as shown in Table \ref{tab:tab1_coco5k}. 
        Following the common protocol \cite{zhang2020context,chen2020imram,wei2020multi}, SMFEA achieves 4.2\%, 8.8\%, and 8.9\% improvements in terms of rSum compared with the state-of-the-art methods CAAN \cite{zhang2020context}, IMRAM \cite{chen2020imram} and MMCA \cite{wei2020multi}, respectively. 
        Especially on the larger test set, the proposed SMFEA model clearly  demonstrates its strong effectiveness with the huge improvements.

\begin{table}[t]
\scriptsize
%\vspace{-0.5em}
\begin{center}
\fontsize{8.5}{12}\selectfont
\renewcommand\tabcolsep{5.0pt}
% \vspace{-0.5em}
\caption{Comparison results on cross-dataset generalization from MS-COCO to Flickr30k.} \label{tab:tab_transfor}
\vspace{-1em}
\begin{tabular}{c|cccc}
\hline

\hline
\multicolumn{1}{c|}{\multirow{2}{*}{Method}} & \multicolumn{2}{c}{Sentence Retrieval} & \multicolumn{2}{c}{Image Retrieval}   \\
\multicolumn{1}{c|}{}                        & R@1       & R@10      & R@1  & \multicolumn{1}{c}{R@10}  \\ \hline

    VSE++ \cite{faghri2017vse++}       & 40.5	& 77.7	& 28.4	& 66.6 \\
    LVSE \cite{engilberge2018finding}	    & 46.5	& 82.2	& 34.9	& 73.5 \\
    SCAN \cite{lee2018stacked}	    & 49.8	& 86.0	& 38.4	& 74.4 \\
    CVSE \cite{wang2020consensus}	    & \underline{56.4}	& \textbf{89.0}	& \underline{39.9}	& \underline{77.2} \\
    \hline 	
    SMFEA(ours)       & \textbf{57.1}  & \underline{88.4}  & \textbf{41.0} & \textbf{80.4} \\ \hline
    
\hline
\end{tabular}
\end{center}
\vspace{-1em}
\end{table}            
\begin{table}[t]
\scriptsize
%\vspace{-0.5em}
\begin{center}
\fontsize{8.5}{12}\selectfont
\renewcommand\tabcolsep{4.0pt}
\caption{Ablation studies on Flickr30K 1K test set.} \label{tab:tab3_ab}
\vspace{-1em}
\begin{tabular}{c|cccccc}
\hline

\hline
\multicolumn{1}{c|}{\multirow{2}{*}{Method}} & \multicolumn{3}{c}{Sentence Retrieval} & \multicolumn{3}{c}{Image Retrieval}   \\
\multicolumn{1}{c|}{}                        & R@1        & R@5       & R@10      & R@1  & R@5 & \multicolumn{1}{c}{R@10}  \\ \hline
    w/o trees       &  71.7          & 91.5          & 94.7          & 51.3          & 78.9          &  87.2           \\
    w/o $\mathcal{D}_{KL}$        & 72.1           &  90.9          &  94.3      & 53.2      & 81.1          & 86.6             \\
    w/o $\mathcal{L}_{CE}$         & 72.4          & 91.4         & 94.9          & 53.8          & 81.5         & 87.0           \\ 
    \hline
    
    SMFEA       & \textbf{73.7}  & \textbf{92.5} & \textbf{96.1}  & \textbf{54.7} & \textbf{82.1} & \textbf{88.4}
    \\ \hline
    
\hline
\end{tabular}
\end{center}
\vspace{-1em}
\end{table}        

\begin{table}[t]
\scriptsize
%\vspace{-0.5em}
\begin{center}
\fontsize{8.5}{12}\selectfont
\renewcommand\tabcolsep{4.0pt}
\caption{Effects of different configurations of hyperparameters $\beta_*$ on Flickr30K 1K test set.} \label{tab:tab3_lambda}
\vspace{-1em}
\begin{tabular}{c|cccccc}
\hline

\hline
\multicolumn{1}{c|}{\multirow{2}{*}{$[\beta_d, \beta_t, \beta_c]$}} & \multicolumn{3}{c}{Sentence Retrieval} & \multicolumn{3}{c}{Image Retrieval}   \\
\multicolumn{1}{c|}{}                        & R@1        & R@5       & R@10      & R@1  & R@5 & \multicolumn{1}{c}{R@10}  \\ \hline
    $[1.0,0.0,0.0]$       &  64.1         &  88.6          &     92.6       &  47.3        &  75.2         &  84.1            \\
    $[0.6,0.0,0.4]$       &  66.5          & 89.9          & 93.7          &     49.1        & 76.9          &  85.0           \\
    $[0.6,0.4,0.0]$       &  70.9          & 90.8          & 93.7          &     52.1        & 79.3          &  86.9           \\
    \hline
    SMFEA       & \textbf{73.7}  & \textbf{92.5} & \textbf{96.1}  & \textbf{54.7} & \textbf{82.1} & \textbf{88.4} \\ \hline
    
\hline
\end{tabular}
\end{center}
\vspace{-0.5em}
\end{table} 
        \subsubsection{Generalization ability for domain adaptation} 
        In order to further verify the generalization of our proposed SMFEA, we conduct the challenging cross-dataset generalization ability experiments which are meaningful for evaluating the cross-modal retrieval performance in real-scenario. 
        Particularly, similar to CVSE \cite{wang2020consensus}, we transfer our model trained on MS-COCO to Flickr30K dataset. 
        As shown in Table \ref{tab:tab_transfor}, our SMFEA achieves significantly outperforms the baseline CVSE \cite{wang2020consensus}, especially in terms of R@1 for both modalities retrieval. 
        It reflects that SMFEA is highly effective and robust for image-sentence retrieval with excellent capability of generalization.

    \subsection{Ablation Studies}
    We perform detailed ablation studies on Flickr30K to investigate the effectiveness of each component of our SMFEA. 
    
    \subsubsection{Effects of different configurations of context-aware tree encoders} 
\begin{figure}[t] %%%%%%%%%%%%%%%%%fig4
	\centering
	\vspace{-0.2em}
	\includegraphics[width=1\linewidth]{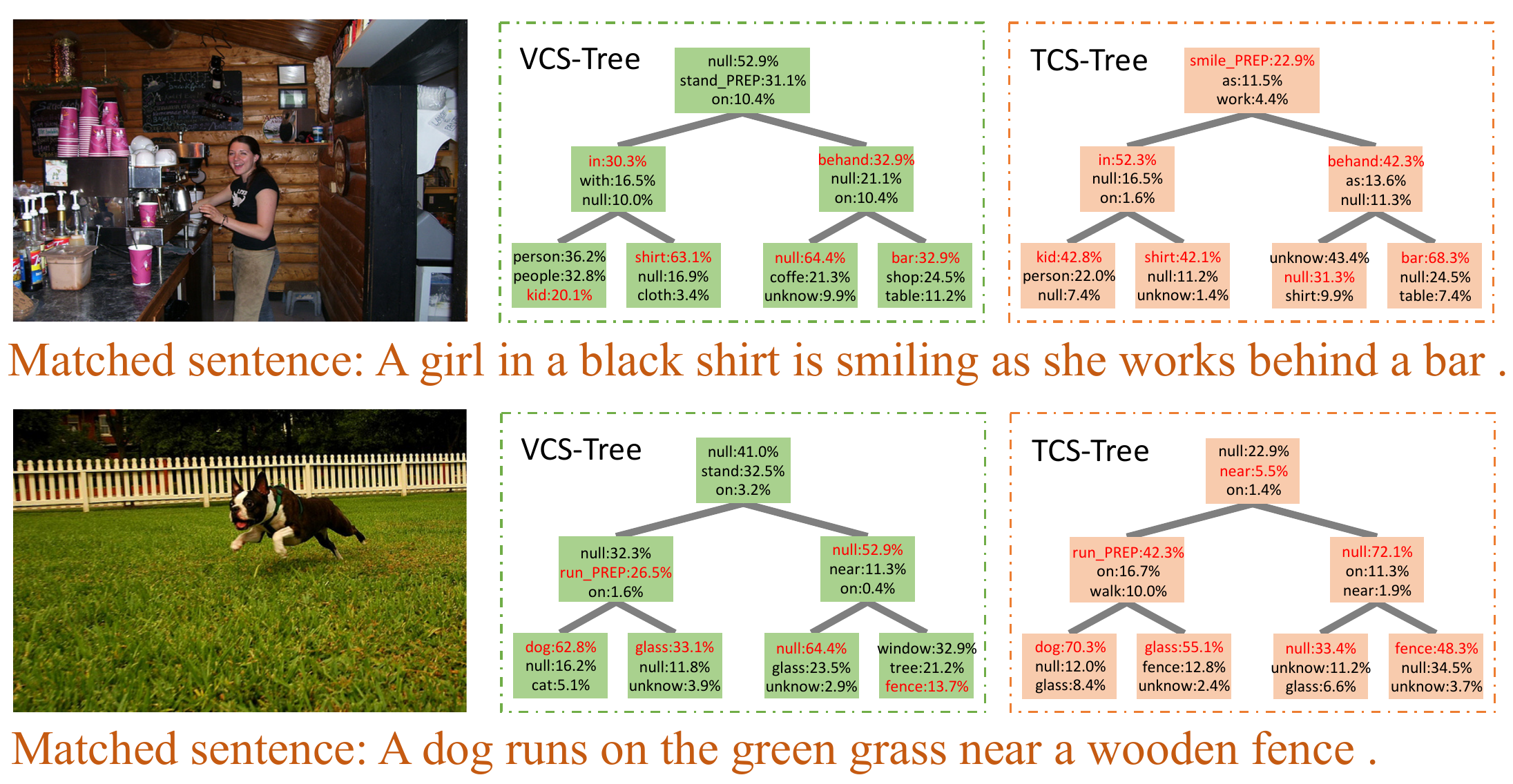}
	\vspace{-2em}
	\caption{Visualization of learned VCS-Tree and TCS-Tree in our SMFEA on Flickr30K. The red font means the correct semantic items according to the referral tree (best viewed in color).}
	\label{fig:example_trees}
	\vspace{-1em}
\end{figure}

    Table \ref{tab:tab3_ab} shows the comparing between SMFEA and its corresponding baselines.
    SMFEA decreases absolutely by 2.0\% and 3.4\% in terms of R@1 for sentence and image retrieval on Flickr30K when removing the multi-modal context-aware structure tree encoders (indicated by w/o trees in Table \ref{tab:tab3_ab}). 
    More detailed, comparison shows that removing $\mathcal{D}_{KL}$ or $\mathcal{L}_{CE}$ makes absolute 3.1\% and 2.2\% drop in terms of R@1-Sum (summing R@1 for image retrieval and sentence retrieval) on Flickr30K, respectively. 
    It has shown that the context-aware structure tree encoders with joint $\mathcal{D}_{KL}$ or $\mathcal{L}_{CE}$ objectives can slightly improve the effectiveness. 
    Please note that our SMFEA without tree encoders (indicated by w/o trees) is reproduced by using the official codes of CVSE \cite{zhang2020context} with slightly different parameters, which may result in different performances compared with \cite{zhang2020context}. 
    In addition, to better understand how the proposed SMFEA model learns the cross-modal fragments/relations, we visualize the learned relation and fragment categories of nodes in VCS-Tree and TCS-Tree in Figure \ref{fig:example_trees}. 
    The proposed VCS-Tree and TCS-Tree capture the intrinsic context semantic relation among the fragments in image and sentences in the in-order traversal manner. 
    Also, the explicit consistency of the inter-modal corresponding tree nodes is fully excavated. 
    %, which suggests that our multi-modal context-aware structured trees can mine the semantic and structure of both modalities. homogeneous

\begin{table}[t]
\scriptsize
%\vspace{-0.5em}
\begin{center}
\fontsize{8.5}{12}\selectfont
\renewcommand\tabcolsep{4.0pt}
\caption{Effects of different encoding structures of SMFEA on Flickr30K 1K test set.} \label{tab:tree_type}
\vspace{-1em}
\begin{tabular}{c|cccccc}
\hline

\hline
\multicolumn{1}{c|}{\multirow{2}{*}{Module}} & \multicolumn{3}{c}{Sentence Retrieval} & \multicolumn{3}{c}{Image Retrieval}   \\
\multicolumn{1}{c|}{}                        & R@1        & R@5       & R@10      & R@1  & R@5 & \multicolumn{1}{c}{R@10}  \\ \hline
    chain-based       &  70.7         &  91.4          &     95.6       &  51.2        &  80.4        &  86.7            \\
    linear-based       &  71.2          & 91.8          &  95.3          &     51.7        & 81.0          &  87.2           \\
    \hline
    SMFEA       & \textbf{73.7}  & \textbf{92.5} & \textbf{96.1}  & \textbf{54.7} & \textbf{82.1} & \textbf{88.4} \\ \hline
    
\hline
\end{tabular}
\end{center}
\vspace{-1em}
\end{table}

    \subsubsection{Effects of different embedding structures of SMFEA} 
    As shown in Table \ref{tab:tree_type}, SMFEA decreases absolutely 1.92\% in terms of the average of all metrics on Flickr30k when replacing context-aware tree structure by a chain-based approach \cite{hochreiter1997long}. 
    In addition, the linear-based tree \cite{chen2017structcap} degrades the average score by 1.55\% compared with our SMFEA. 
    These observations suggest that our context-aware tree encoders can improve the semantic and structural context consistency mining effectiveness between visual and textual features.
% \begin{figure}[t] %%%%%%%%%%%%%%%%%fig4
% 	\centering
% 	%\vspace{-0.4em}
% 	\includegraphics[width=0.9\linewidth]{Figure_1.eps}
% 	\vspace{-1.5em}
% 	\caption{Comparisons of the average number of labeled matching sentences in the top 5 ranked search results (the Avg. LT5) on different datasets (best viewed in color).}
% 	\label{fig:AvgGTSentence}
% 	\vspace{-0.5em}
% \end{figure}

    \subsubsection{Effects of different configurations of hyperparameters $\beta_*$} 
    We evaluate the impact of different multi-modal representations in Eq.(\ref{eq:VF}) and Eq. (\ref{eq:SF}), including the instance-level features ($V^D/S^D$), consensus-aware concepts representations ($V^C/S^C$) and context-aware structured tree embedding and aligning features ($V^T$ $/S^T$), for image-sentence retrieval. 
    As shown in Table \ref{tab:tab3_lambda}, $[\beta_d, \beta_t,$ $ \beta_c]$ denotes different balance parameters in Eq.(\ref{eq:VF}). For instance, $\beta_d$ denotes the proportion of SMFEA employing the instance-level multi-modal features. 
    Combining all three representations ($[\beta_d, \beta_c,$ $\beta_t]=[0.6, 0.2, 0.2]$) in SMFEA achieves the best performance over all metrics. 
    Moreover, compared with combining the CAC ($[\beta_d, \beta_t, \beta_c]=[0.6, 0.0, 0.4]$), combining the multi-modal context-aware structured tree features with alignment model ($[\beta_d, \beta_t, \beta_c]=[0.6, 0.4, 0.0]$) achieves 12.6\% improvement in terms of rSum on Flickr30.  
    It is obvious that our multi-modal context-aware structured tree embedding and alignment model improves the larger performance boost for both modalities retrieval, which validates that the importance of learning the intra-modal relations and inter-modal consistence of tree nodes correspondence.

\begin{figure}[t] %%%%%%%%%%%%%%%%%fig2
	\centering
	\vspace{0.3em}
	\includegraphics[width=1.0\linewidth]{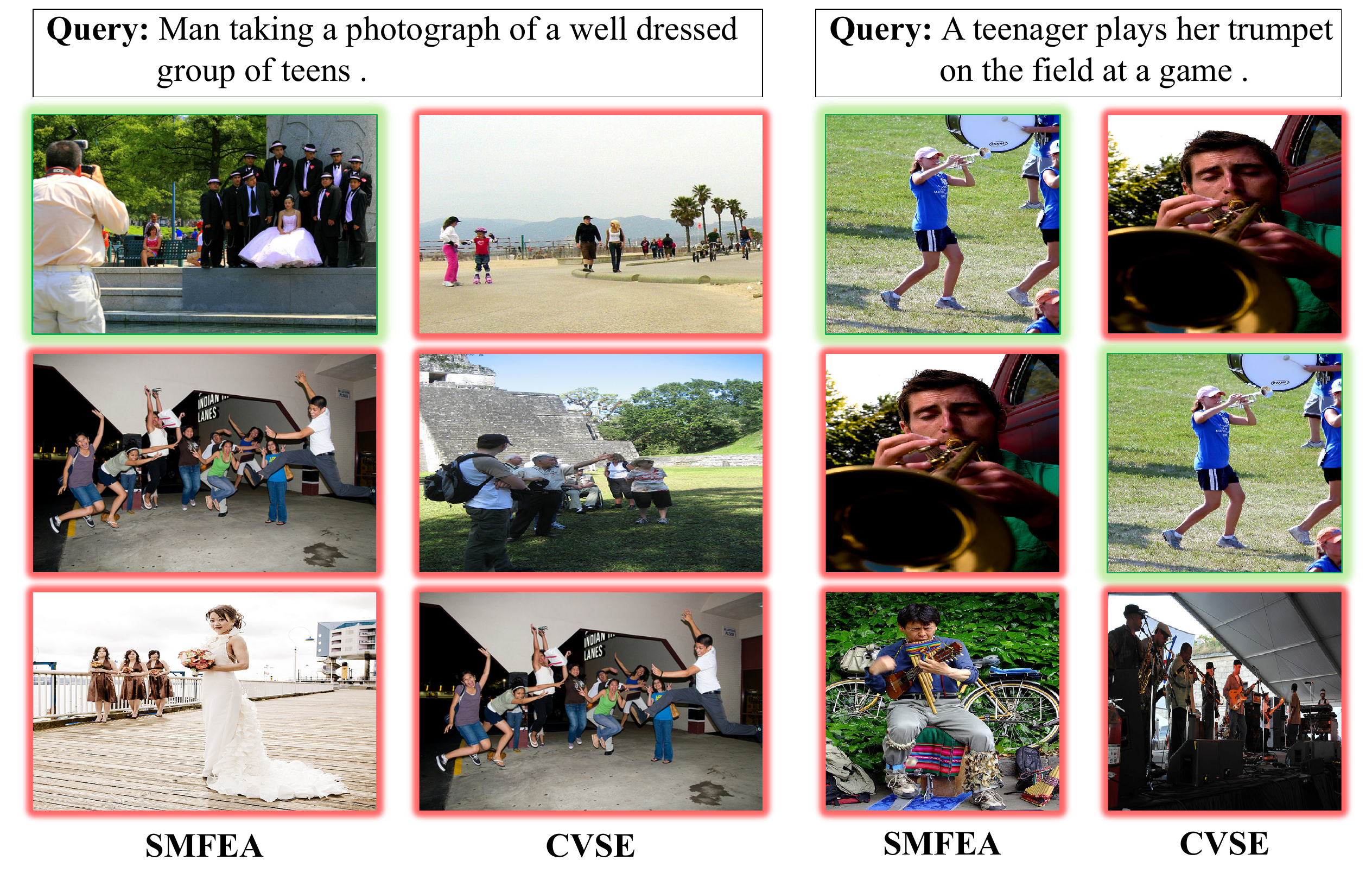}
	\vspace{-2em}
	\caption{Visual comparisons of image retrieval between our SMFEA and CVSE \cite{wang2020consensus} on Flickr30K (best viewed in color).} %For each sentence query, we present the top 3 ranked images, ranking from top to bottom. The correctly matched images are marked in green and the mismatched images are marked in red (best viewed in color).}
	\label{fig:example_t2i}
	%\vspace{-1em}
\end{figure}
\begin{figure}[t] %%%%%%%%%%%%%%%%%fig3
	\centering
	\vspace{-0.6em}
	\includegraphics[width=1\linewidth]{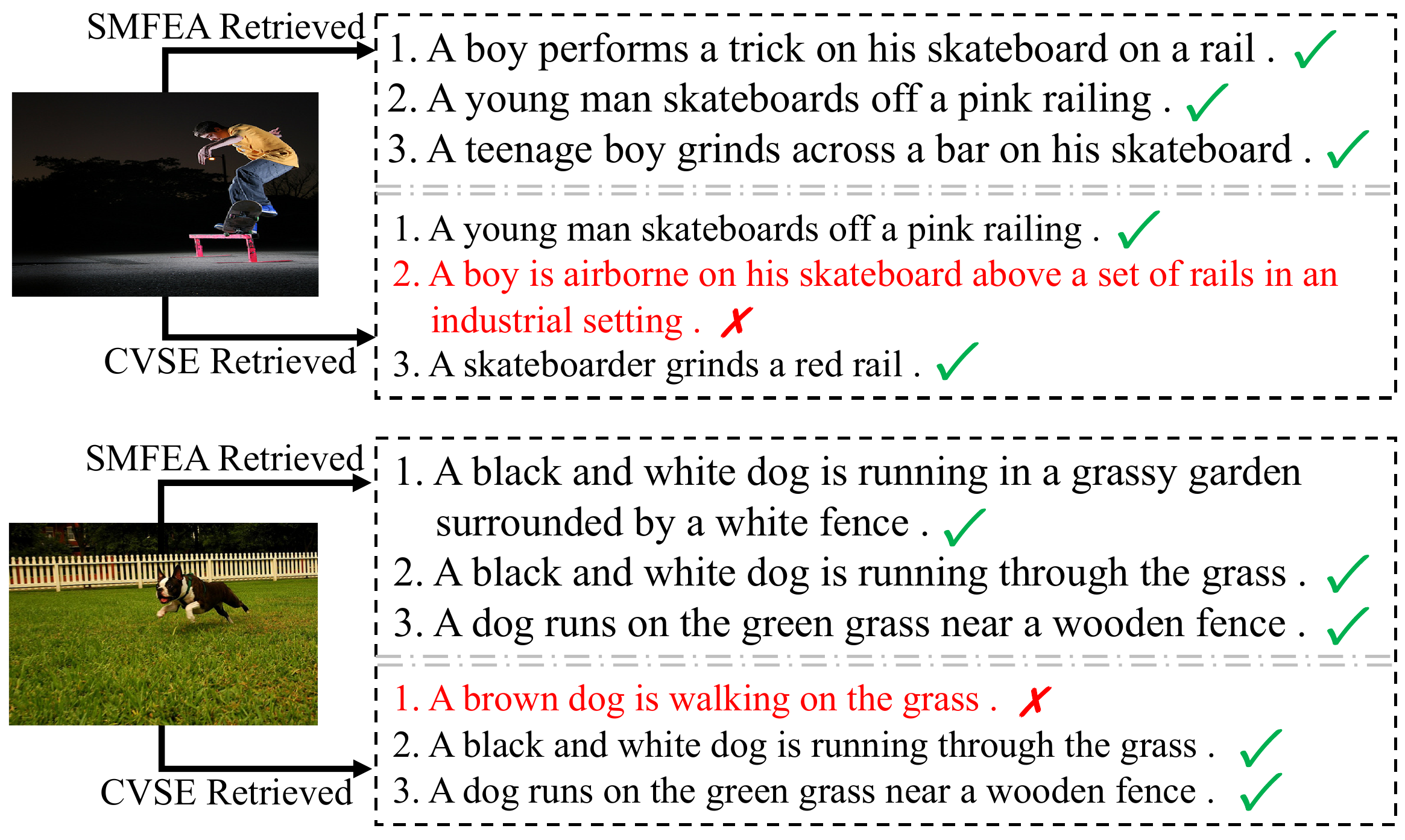}
	\vspace{-1.5em}
	\caption{Visual comparisons of sentence retrieval examples between SMFEA and CVSE \cite{wang2020consensus} on Flickr30K (best viewed in color).} %. The mismatched texts are marked as red 
	\label{fig:example_i2t}
	\vspace{-1em}
\end{figure} 
        \subsection{Visualization of results} 
\begin{figure}[t] %%%%%%%%%%%%%%%%%fig3
	\centering
	\vspace{0.3em}
	\includegraphics[width=1\linewidth]{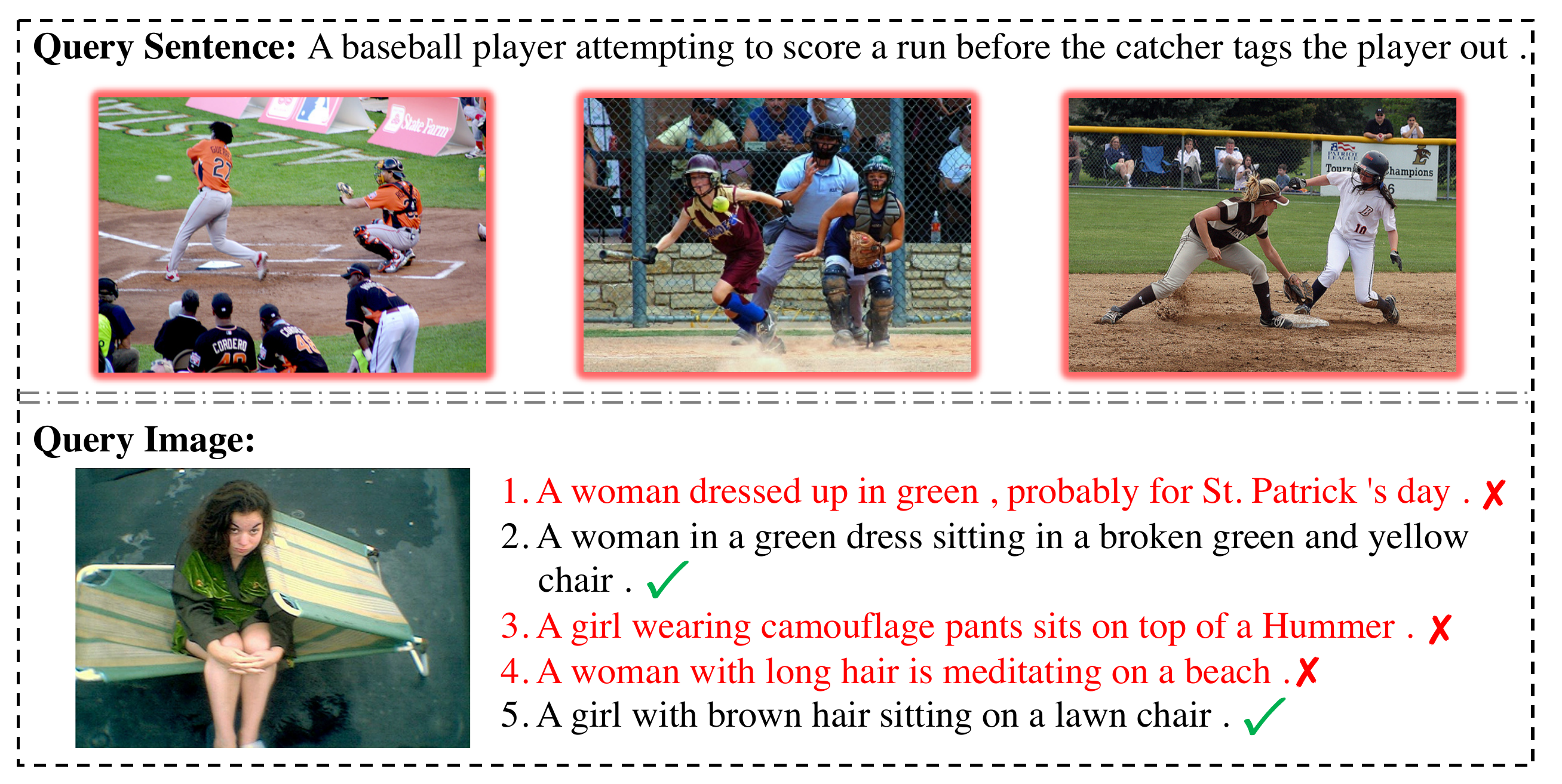}
	\vspace{-2em}
	\caption{Visualization of the failed image retrieval and sentence retrieval examples on Flickr30K by SMFEA (best viewed in color).} %. The mismatched texts are marked as red 
	\label{fig:badcases}
	\vspace{-1em}
\end{figure}

        To better understand the effectiveness of our proposed model, we visualize matching results of the sentence retrieval and image retrieval on Flickr30K in Figure \ref{fig:example_t2i} and Figure \ref{fig:example_i2t}. 
        For image retrieval shown in Figure \ref{fig:example_t2i}, we show the top 3 ranked images for each text query matched by our proposed SMFEA in first column, and followed by CVSE \cite{wang2020consensus} in the second column. 
        The true matches are outlined in green boxes and false matches in red. 
        Furthermore, as shown in Figure \ref{fig:example_i2t}, we visualize the sentence retrieval results (top-3 retrieved sentences) predicted by SMFEA  and CVSE \cite{wang2020consensus}, where the mismatches are highlight in red. 
        Examples of failed image retrieval and sentence retrieval are shown in Figure \ref{fig:badcases}. 
        However, in this case the wrong images/sentences have   similar semantic or structural content to true matches. 
        We argue that the reason for this phenomenon may be that our current tree structure model is to unify the coarse-grained semantics and structural consistency between the two modalities. 
        It has a good ability to improve the robustness of the model.
        But there are certain shortcomings in the distinction of similar sets.
        We will build fine-grained vocabularies to improve future work.

\section{Conclusion and Future work}
In this paper, we exploit image-sentence retrieval with structured multi-modal feature embedding and cross-modal alignment. 
Our work serves as the first to narrow the cross-modal heterogeneous gap by aligning the explicitly inter-modal semantic and structure correspondence between images and sentences with the visual/textual inner context-aware structured tree encoder (VCS-Tree/TCS-Tree) capturing. 
We proposed a novel structured multi-modal feature embedding and alignment (SMFEA) model, which contains a VCS-Tree and a TCS-Tree to enhance the intrinsic context-aware structured semantic information for image and sentence, respectively.  
Furthermore, the consistency estimation of the corresponding inter-modal tree nodes is maximized to narrow the cross-modal pair-wise distance. 
Extensive quantitative comparisons demonstrate that our SMFEA can achieve state-of-the-art performance across popular standard benchmarks, MS-COCO and Flickr30K, under various evaluation metrics.

Future work includes the exploration of fine-grained category expansion of fragment/relation in the context-aware structured tree encoders, improving the accuracy and fine-grained representation of the referral tree, and so on.

\begin{acks}
This work was supported by National Key R\&D Program of China, under Grant No. 2020AAA0104500.
\end{acks}
%\end{verbatim} 

\bibliographystyle{ACM-Reference-Format}
\balance
\bibliography{sample-base}

\end{document}